\documentclass[conference]{IEEEtran}
\IEEEoverridecommandlockouts
\usepackage{cite}
\usepackage{amsmath,amssymb,amsfonts}
\usepackage{algorithmic}
\usepackage{graphicx}
\usepackage{textcomp}
\usepackage{xcolor}
\usepackage{mathtools}
\usepackage{hyperref}
\def\BibTeX{{\rm B\kern-.05em{\sc i\kern-.025em b}\kern-.08em
    T\kern-.1667em\lower.7ex\hbox{E}\kern-.125emX}}

\begin{document}
\thispagestyle{empty} 
\onecolumn

\begin{changemargin}{3cm}{3cm} 
    \vspace*{\fill}
    \section*{Copyright Notice}
        \begin{center}
            © 2022 IEEE.  Personal use of this material is permitted.  Permission from IEEE must be obtained for all other uses, in any current or future media, including reprinting/republishing this material for advertising or promotional purposes, creating new collective works, for resale or redistribution to servers or lists, or reuse of any copyrighted component of this work in other works.

            \vspace{3mm}

            The final version is available at: \href{https://doi.org/10.1109/IECON49645.2022.9968678}{https://doi.org/10.1109/IECON49645.2022.9968678}
            
        \end{center}
        
    \vspace*{\fill}

\end{changemargin} 

{
	\let\clearpage\relax
}

\twocolumn

\title{Safety Aware Autonomous Path Planning Using Model Predictive Reinforcement Learning for Inland Waterways
}

\author{
    \IEEEauthorblockN{Astrid Vanneste\IEEEauthorrefmark{1}, 
                      Simon Vanneste\IEEEauthorrefmark{1}, 
                      Olivier Vasseur\IEEEauthorrefmark{1}, 
                      Robin Janssens\IEEEauthorrefmark{1},
                      Mattias Billast\IEEEauthorrefmark{2},\\
                      Ali Anwar\IEEEauthorrefmark{1},
                      Kevin Mets\IEEEauthorrefmark{2},
                      Tom De Schepper\IEEEauthorrefmark{2},
                      Siegfried Mercelis\IEEEauthorrefmark{1},
                      Peter Hellinckx\IEEEauthorrefmark{1}
                      }
    \IEEEauthorblockA{\textit{IDLab - \IEEEauthorrefmark{1}Faculty of Applied Engineering, \IEEEauthorrefmark{2}Department of Computer Science} \\
    \textit{University of Antwerp - imec}\\
    Antwerp, Belgium \\
    \{astrid.vanneste, simon.vanneste, olivier.vasseur, robin.janssens, mattias.billast, ali.anwar,\\kevin.mets, tom.deschepper, siegfried.mercelis, peter.hellinckx\}@uantwerpen.be
}
}

\maketitle

\begin{abstract}
In recent years, interest in autonomous shipping in urban waterways has increased significantly due to the trend of keeping cars and trucks out of city centers.
Classical approaches such as Frenet frame based planning and potential field navigation often require tuning of many configuration parameters and sometimes even require a different configuration depending on the situation. In this paper, we propose a novel path planning approach based on reinforcement learning called Model Predictive Reinforcement Learning (MPRL). MPRL calculates a series of waypoints for the vessel to follow. The environment is represented as an occupancy grid map, allowing us to deal with any shape of waterway and any number and shape of obstacles. We demonstrate our approach on two scenarios and compare the resulting path with path planning using a Frenet frame and path planning based on a proximal policy optimization (PPO) agent. Our results show that MPRL outperforms both baselines in both test scenarios. The PPO based approach was not able to reach the goal in either scenario while the Frenet frame approach failed in the scenario consisting of a corner with obstacles. MPRL was able to safely (collision free) navigate to the goal in both of the test scenarios. 

\end{abstract}

\begin{IEEEkeywords}
reinforcement learning, autonomous path planning, collision avoidance, autonomous shipping
\end{IEEEkeywords}

\section{Introduction}
In recent years, more and more cities such as London, Antwerp, Berlin, etc. have introduced low emission zones or reduced the allowed traffic in the city centre \cite{LEZ2020}. However, goods still need to be transported within these cities. In cities with dense waterways (e.g. Ghent, Bruges, Göteborg, Stockholm, Lyon, Berlin, Hamburg) traffic can be significantly reduced by moving Last Mile Logistics to the waterways. In Amsterdam and Utrecht (manually steered) urban shipping is already in use, but broad applicability is hindered by the cost and shortage of personnel. By using autonomous navigation systems, this cost can be reduced, allowing for a wider use of waterways. In this paper, we propose a path planning system designed for use in inland waterways based on reinforcement learning (RL) called Model Predictive Reinforcement Learning (MPRL) which can be trained to navigate the waterway instead of using manually engineered heuristics. To train and test our system we designed a novel simulation environment. 
We use this simulation environment to compare our method to Frenet frame \cite{frenet} navigation and to proximal policy optimization (PPO) \cite{schulman2017proximal} based navigation. Frenet frame control\cite{frenet} is a well established control algorithm that has been applied on various control applications. PPO \cite{schulman2017proximal} is an actor-critic based RL algorithm that has shown state-of-the-art performance in many RL benchmarks. 
Furthermore, our path planning system can be used as an assistance system for a skipper who remotely monitors multiple ships. The output consists of waypoints that can be followed by an automatic control system on the vessel. Whenever the vessel encounters a situation that cannot be handled by our path planning algorithm, the system turns the control over to the skipper.
Our main contributions consist of presenting a complete navigation system for autonomous shipping, consisting of global and local navigation and including a failure mode. To evaluate our approach we present a simulation system and compare our approach with two baselines. 

The remainder of this paper is structured as follows. Section \ref{sec:related_work} investigates related work. Section \ref{sec:methods} explains our path planning system and the baselines. We discuss the details of our experiments in Section \ref{sec:experiments}. In Section \ref{sec:results} we look at the experimental results of our method. In Section \ref{sec:conclusion} we draw our conclusions from these experiments. 

\section{Related Work}
\label{sec:related_work}

In this section, we review the state of the art relevant to our research. We start with classical methods and then discuss methods that use RL based path planning.

In general, classical path planning approaches are highly dependent on different mission scenarios and therefore challenging to use as a generic path planning method.
Dijkstra’s algorithm \cite{wang2011application} utilizes a grid map to find the path to the goal before any movement. The algorithm searches the route between two positions by examining the neighbors of a parent node during each iteration.
The A* algorithm \cite{astar} is a local path planning algorithm that adds a heuristic function to Dijkstra’s algorithm. 
M. Seder et al. \cite{seder2007dynamic} integrated the D* search algorithm and the dynamic window approach (DWA) to plan the agent’s path  based on the kinodynamic requirements. DWA samples multiple velocities  and provokes a series of intrinsic motion trajectories in a specific time. By comparing the scores of various trajectories, the algorithm would determine the optimal trajectory for the agent.
Artificial potential field (APF) path planning \cite{Khatib1990apf} is a path planning method that uses different potential forces. The destination applies an attractive force and obstacles apply repulsive forces. The sum of these forces forms a potential field. APF path planning constructs a path by following the force field formed by the potential field. A downside to this approach is that it is prone to getting stuck in local optima in the APF. 
Model Predictive Control (MPC) \cite{garcia1989model} is a control method which can iteratively optimize a set of parameters while taking into account future events by using a dynamics model of the process. The process is optimized in order to reduce a certain cost function which evaluates the process horizon. However, MPC requires a well tuned, manually engineered cost function in order to select the correct set of control parameters.
A big disadvantage of many classical approaches is that they require tedious tuning of many application specific parameters. In RL this problem is less apparent.
Patel et al.\cite{patel2020dynamically} presented a hybrid DWA-RL motion planning approach. The planner employs the RL algorithm as the top-level policy optimizer and adopts DWA as the low-level observation space generator. DWA-RL benefits from DWA to perform kinodynamically feasible planning and uses RL to select the optimal velocity commands to maximize the global returns for complex environments. Lu Chang et al.\cite{chang2021reinforcement} proposed Q-learning-based DWA. Q-learning-based DWA uses a Q-learning RL module to auto-tune the weights in the DWA evaluation function at each timestep to improve the optimality of the planner.

Zhang et al.\cite{zhang2019} use deep Q-learning \cite{dqn} to control a vessel. As an improvement to this approach they present a version where they use APF combined with the deep Q-learning approach which improved their results significantly. 
Shen et al. \cite{shen2019} focus on collision avoidance of multiple ships by using deep Q-learning. 
Zhao and Roh \cite{zhao2019} propose an RL method that focuses on compliance with the Convention on the International Regulations for Preventing Collisions at Sea (COLREGs). They train a proximal policy optimization (PPO)\cite{schulman2017proximal} agent that controls the rudder by choosing one of three discrete rudder actions.
Guo et al.\cite{guo2020} propose a system that uses Deep Deterministic Policy Gradient (DDPG)\cite{ddpg} to control the rudder and acceleration of the vessel. They also propose an approach that uses APF with DDPG to control the vessel. In their work, they focus on navigation in open water while our focus is on navigation in narrow waterways.
Each of these RL methods chooses to directly control the heading and/or speed of the ship. Our method, on the contrary, has waypoints as output. We choose to use waypoints since this allows the skipper to conveniently monitor the behaviour. This choice does mean that we need an additional path following algorithm to follow these waypoints. Autonomy is often defined using different levels \cite{autonomylevels}. We aim for level 3 autonomy. Level 3 autonomy means that the ship sails with human supervision allowing human intervention when necessary. The sequence of waypoints allows the skipper to evaluate whether the ship will enter a dangerous situation where they need to intervene. Performing similar monitoring while directly controlling the rudder and thrust is challenging and impractical.
Another key difference between our method and the state of the art is that we choose to use an occupancy grid map to represent the environment and obstacles. This allows us to sail in small unstructured waterways with an unknown number of obstacles.

\begin{figure}[t]
    \centering
    \includegraphics[width=0.75\linewidth]{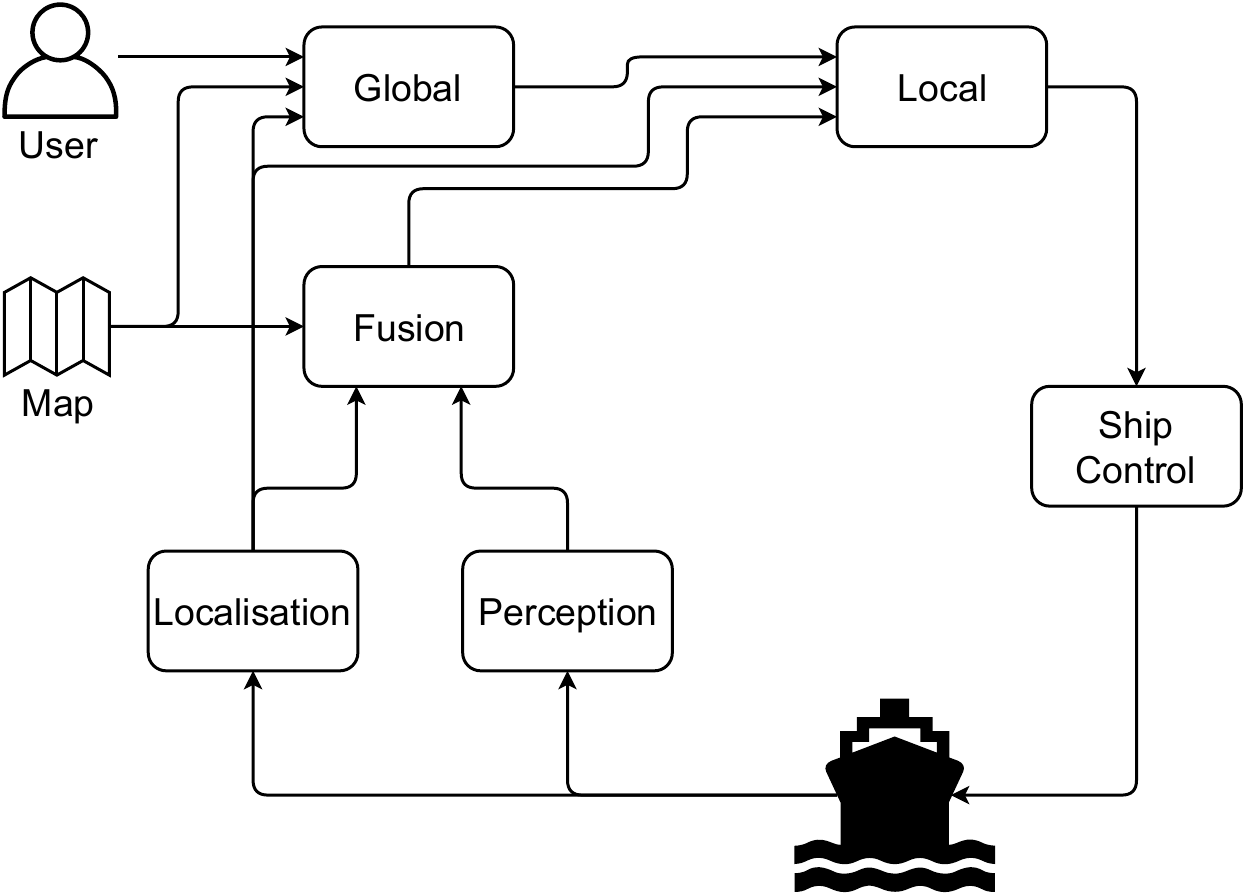}
    \caption{Architecture of the path planning system in which the user sets a goal location and the system plans a trajectory using a map of the environment along with sensor and localisation information}
    \label{fig:architecture}
\end{figure}

\section{Methods}
\label{sec:methods}
In this section, we explain the different aspects of our path planning system. 

\subsection{Architecture}
In order to create our autonomous navigation system, we need to combine multiple elements of input data to provide the local path planner with the correct information. Figure \ref{fig:architecture} shows the architecture of our system. Before we start navigating, the user provides the system with the end goal that we need to reach. Based on our current position, the end goal and information from OpenStreetMaps \cite{osm}, we calculate a global trajectory from the current position to the end goal. This global trajectory provides the local planner with intermediate goals. Our local planner also requires information about the current state the ship is in. Both static and dynamic information about the environment are fused into one occupancy grid map that is provided to the local planner. This makes sure that both dynamic information provided by sensors and the static information obtained from OpenStreetMaps \cite{osm} are being treated in the same way by the planner. The local path planner generates a local trajectory that is passed on to the ship control, which controls the rudder and thrust of the ship to follow the provided trajectory. This architecture allows us to compare the proposed RL planner with the Frenet frame baseline and the PPO baseline by switching the local planner between these algorithms while keeping the other interfaces the same. This paper focuses on the path planning aspect of autonomous navigation. Therefore, the ship control is simulated and the sensor inputs are removed so that all the obstacle information is included in the static map.

\subsection{Simulation}
\label{sec:simulation_env}
In order to safely and efficiently test our system as well as train the models for the RL based path planning, we needed a simulation system. Since it will be used for the RL based path planning, we chose to use the multiagent RLlib \cite{rllib} extension of the gym interface by OpenAI \cite{openai_gym}. The dynamics of the ship are based on the uSimMarine dynamics included in MOOS-ivp \cite{moos_ivp}. In addition, we also include a drag component on the speed of the vessel using the drag equation in Eq. (\ref{eq:drag}) \cite{drag_rayleigh1876liii}. Here, $F_d$ represents the drag force, $\rho$ represents the mass density of the fluid, $u$ is the flow velocity relative to the object, $A$ is the reference area and $c_{\mathrm{d}}$ is the drag coefficient. 

\begin{equation}
    F_{\mathrm{d}} = \frac{1}{2} \rho u^2 c_{\mathrm{d}} A
    \label{eq:drag}
\end{equation}

To be able to control the vessel using a desired speed and heading value we add two PID controllers. One PID controller controls the thrust of the ship to achieve the desired speed. The second PID controller controls the rudder angle to get the desired heading. 

We use binary occupancy grid maps to represent the environment. In this occupancy grid map (an example is shown in Figure \ref{fig:results_scen1}) a zero represents an obstacle and a one represents a part of the waterway. Collisions can be efficiently detected by taking a subsection of the occupancy grid map which the ship passes during a transition in a straight line. 
If this subsection contains an obstacle, the transition results in a collision. 

\subsection{Global Path Planning}
\label{subsec:global_path_planning}

The Global Path planning generates waypoints that the local path planning can follow in order to go towards the final goal position. The global path planning uses the public waterway data set of OpenStreetMaps \cite{osm} which can be queried using the Overpass API \cite{overpass}. OpenStreetMaps is able to provide nodes placed in the center of the navigable waterways using the $way[waterway=river]$ and $way[waterway=canal]$ features within a window of coordinates \cite{osm}. These nodes can be connected as a graph structure using the Euclidean distance between the nodes as the link cost. After we have a node graph representation of the waterway system we need to search for the shortest path between these nodes. Our current implementation incorporates Dijkstra's algorithm \cite{dijkstra}. When using bigger, more complex environments the use of a better search heuristic is preferred (e.g.: A* algorithm \cite{astar}). This graph traversal yields an ordered list of nodes leading from the start point (i.e. your current location) to the target goal (i.e. the destination). This ordered list of nodes is processed to achieve the final waypoints which are sent to the local path planning.

\subsection{Proximal Policy Optimization}
\label{sec:PPO}
The simulation environment provides observations consisting of the relative heading of the ship to the target location, the distance of the ship to the target location, the speed of the ship and a square subsection of the environment map around the ship. The simulation can be controlled by using actions consisting of a desired speed and a desired change in heading. The reward is composed of multiple components as can be seen in Eq. (\ref{eq:reward_function}). The first part, explained in Eq. (\ref{eq:reward_function_distance}) is based on the distance of the current position to the goal position. The distance $d$ is normalized using the maximum distance we can start from the goal position, $d_{\mathrm{max}}$. A higher distance from the goal decreases the reward value. Eq. (\ref{eq:reward_function_goal}) describes a bonus reward when getting within a certain distance $D_G$ from the goal. When a collision occurs, a penalty is applied as shown in Eq. (\ref{eq:reward_function_collision}). Lastly, we apply a penalty to the agent for high values for the heading action $a_{\mathrm{h}}$ weighted by $w_{\mathrm{h}}$. The heading action describes the desired change in heading relative to the current heading. This penalty will therefore encourage the agent to gradually change the heading of the ship with a low value for $a_{\mathrm{h}}$ instead of taking sharp turns using high values for $a_{\mathrm{h}}$.

\begin{equation}
    r = r_{\mathrm{d}} + r_{\mathrm{g}} + r_{\mathrm{c}} + r_{\mathrm{h}}
    \label{eq:reward_function}
\end{equation}
\begin{align}
    r_{\mathrm{d}} &= 1 - \frac{d}{d_{\mathrm{max}}} 
    \label{eq:reward_function_distance}\\
    r_{\mathrm{g}} &= 
    \begin{cases}
            r_{\mathrm{goal\_reached}}, \quad  &d < D_{\mathrm{G}}\\
            0, \quad &d \geq D_{\mathrm{G}}
    \end{cases}
    \label{eq:reward_function_goal}\\
    r_{\mathrm{c}} &= 
    \begin{cases}
            r_{\mathrm{collision}}, \quad  &\text{if\ collision}\\
            0, \quad &\mbox{if no\ collision}
    \end{cases}
    \label{eq:reward_function_collision}\\
    r_{\mathrm{h}} &= -w_{\mathrm{h}} . |a_{\mathrm{h}}| 
    \label{eq:reward_function_toggle}
\end{align}

As a baseline, we train a PPO agent \cite{schulman2017proximal} on our environment. We use the policy network to determine a simulation action. We perform several steps of simulation. The resulting new location is used as the output waypoint. We do this multiple times to get the required amount of waypoints. Failure is detected by checking if during the calculation of the first waypoint a collision occurs in the simulation. A downside of this technique is that by performing multiple simulation steps for a single waypoint we cannot be sure that the waypoint can be reached in a straight line without any collisions. This problem is solved in the MPRL method. 

\subsection{Model Predictive Reinforcement Learning}
\begin{figure}[t]
    \centering
    \includegraphics[angle=0,width=0.99\linewidth]{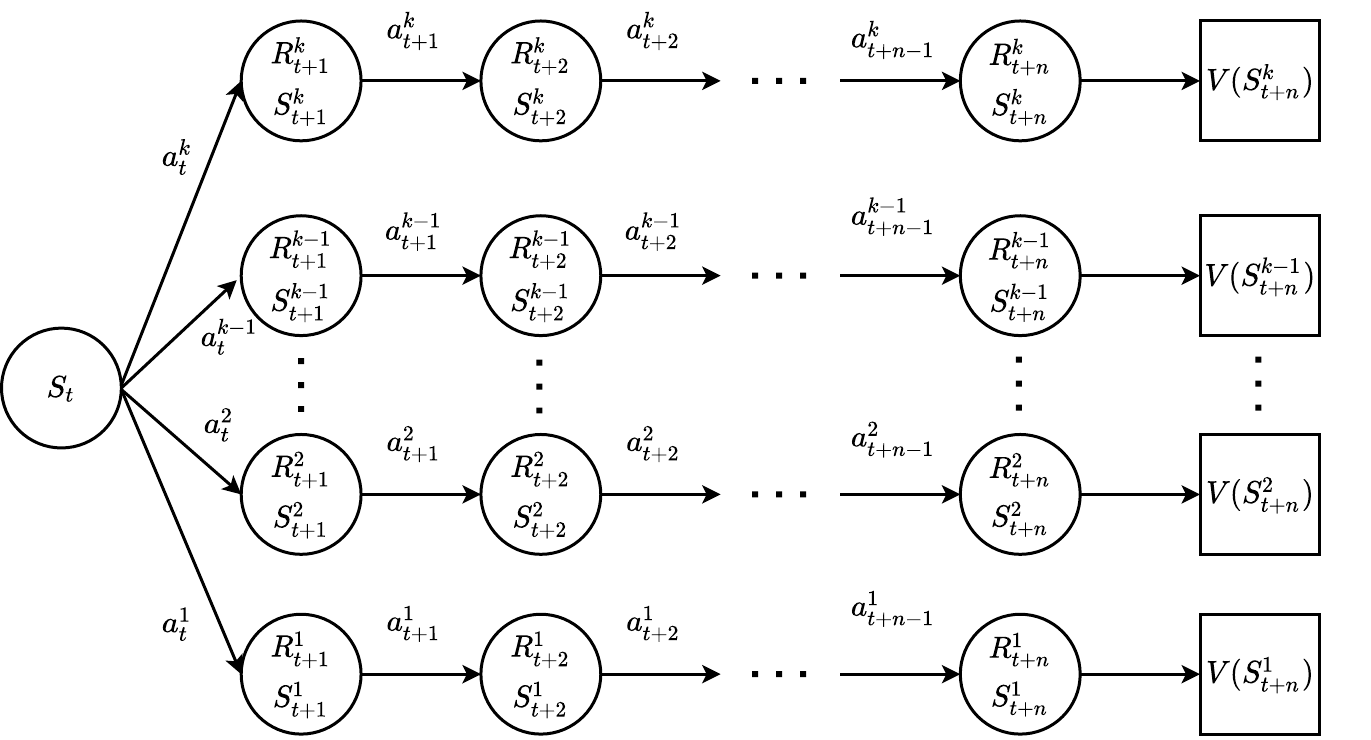}
    \caption{MPRL for $n$ simulated timesteps in $k$ trajectories.}
    \label{fig:n_bootstrapping}
\end{figure}

In our Model Predictive Reinforcement Learning (MPRL) method, we use the PPO agent that we train as described in Section \ref{sec:PPO}. However, for safety concerns and to decouple the local planning from the OEM controllers, the ship cannot be controlled directly but needs to receive waypoints which are translated into motor speed and rudder angle by the ship control algorithm. These waypoints need to be generated by the MPRL algorithm to allow the ship to avoid unforeseen obstacles during operation.
These waypoints are generated by using a combination of the simulator and the PPO agent. First, we generate a number of trajectories the ship can follow using a constant speed. Every trajectory (as shown in Figure \ref{fig:n_bootstrapping}) is defined by the action $a_t$ and the change in action $\Delta a$. These values determine the action sequence used in the trajectory as defined in Eq. (\ref{eq:s_trajectory}). After the second integral, this becomes a third degree polynomial which makes it possible for the ship to generate a variety of different trajectories such as s-turns, regular turns, straight trajectories, etc. As the trajectories are based on $j$ possible actions and $l$ possible changes in action values, we achieve $k = j * l$ trajectories.
\begin{equation}
a_{t+1} = a_t + \Delta a
\label{eq:s_trajectory}
\end{equation}

A certain trajectory is simulated using the simulator in order to obtain the next state and the next reward using the action $a$ which is created using Eq. (\ref{eq:s_trajectory}). This is repeated for $n$ steps to generate the full trajectory. This trajectory is evaluated using the expected return $G_{t:t+n}$ after $n$ simulation steps. The expected return is calculated using n-step bootstrapping \cite{sutton2018reinforcement} as defined by Eq. (\ref{eq:n_bootstrapping}).

\begin{equation}
G_{t:t+n} \doteq R_{t+1} + \gamma R_{t+2} + \dots + \gamma^{n-1} R_{t+n} + \gamma^n V(S_{t+n})
\label{eq:n_bootstrapping}
\end{equation}

After the simulation and evaluation of these trajectories, the trajectory with the highest expected return is selected. This selected trajectory is transformed to waypoints by using the locations achieved in the simulation.

Additionally, this method of generating waypoints allows us to detect failures when generating possible future scenarios. 
The failure detection criterium used in this paper is that the transition to the first waypoint in the trajectory results in a collision.
In this scenario a handover to a human operator is requested to allow for safe operation of the ship. This failure mode can be further extended for specific applications allowing for safe operation.

\begin{figure*}
\begin{minipage}[b]{.99\textwidth}
    \centering
    \includegraphics[width=0.99\linewidth, trim=10 20 0 20, clip]{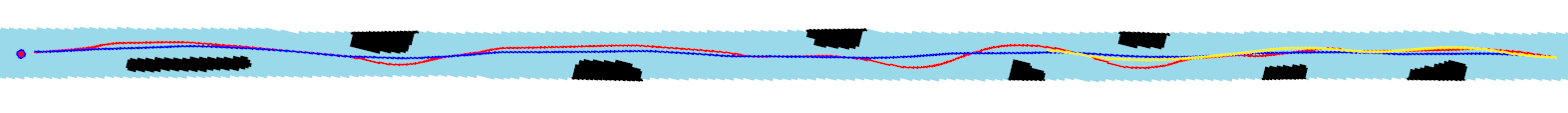}
    \caption{Paths of the different path planning methods in scenario 1 (Red - MPRL, Yellow - PPO, Blue - Frenet Frame Navigation) in which the black pixels (a one in the occupancy grid map) represent the objects on the waterway.}
    \label{fig:results_scen1}
\end{minipage}
\end{figure*}

\begin{figure}[t]
    \centering
    \includegraphics[width=0.75\linewidth, angle=0,origin=c, trim=7 9 7 11, clip]{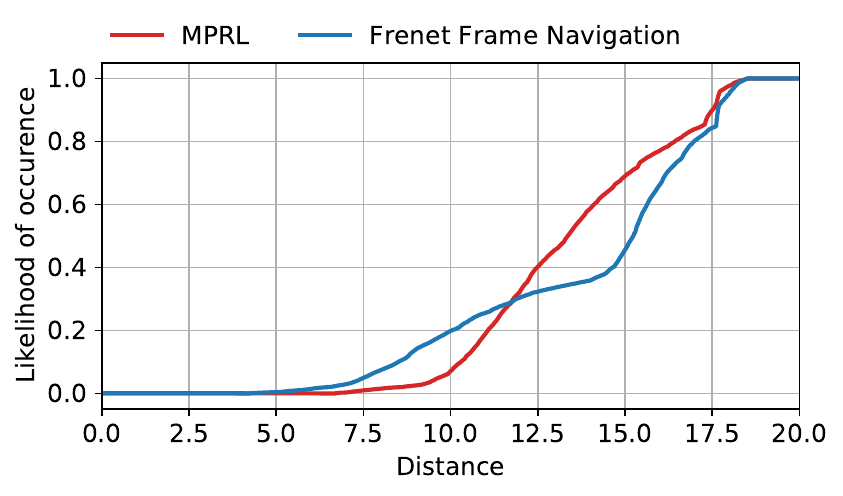}
    \caption{Cumulative distribution function of the probability of the trajectory being a certain distance from the nearest obstacle.}
    \label{fig:CDF_scen1}
\end{figure}

\subsection{Path planning in the Frenet frame}
The path planning method in the Frenet frame assumes that the vessel follows a trajectory that may be offset from a reference trajectory. This reference trajectory is provided by the global path planning presented in section \ref{subsec:global_path_planning}. The Frenet coordinates of the vessel are the arc length $ \sigma $ (longitudinal distance along the reference trajectory) and the distance $ \delta $ (lateral offset from the projection on the reference trajectory).
The Frenet frame method that we have used is explained in detail by Werling et al.\cite{frenet} and our implementation is based on the Optimal Trajectory in a Frenet Frame implementation 
from the PythonRobotics library \cite{python_robotics}.




The algorithm generates trajectories which are checked for validity and rated using a cost function. This cost function is the addition of terms that account for jerk\footnote{The jerk is defined as the third time derivative of each position function $ \sigma $ and $ \delta $.}, time spent by the ship to reach the end position, discrepancies with respect to target parameters and proximity to collisions. More specifically, the total cost reads

\begin{equation}
    C =
    K_{\mathrm{lat}} \,C_{\mathrm{lat}}
    + K_{\mathrm{lon}} \,C_{\mathrm{lon}}
    + K_{\mathrm{col}} \,C_{\mathrm{col}}
    \text{,}
    \label{eq:frenet_total_cost}
\end{equation}

with

\begin{align}
    C_{\mathrm{lat}} &\coloneqq
    K_{\mathrm{J}} \,J(\delta)
    + K_{\mathrm{t}} \,t_{\mathrm{end}}
    + K_{\mathrm{\delta}} \,\delta_{\mathrm{end}}^{2}
    \label{eq:frenet_cost_lat}\\
    C_{\mathrm{lon}} &\coloneqq  
    K_{\mathrm{J}} \,J(\sigma)
    + K_{\mathrm{t}} \,t_{\mathrm{end}}
    + K_{\mathrm{\sigma}} \,(\dot{\sigma}_{\mathrm{end}} - \dot{\sigma}_{\mathrm{target}})^{2}
    \label{eq:frenet_cost_lon}\\
    C_{\mathrm{col}} &\coloneqq
    \sum_{T} \mathrm{exp}(K_{\mathrm{D}} - |D_T|)
    \label{eq:frenet_cost_col}
\end{align}

where the jerk costs $ J $ are defined as 
($ p $ in the below equation can be either $ \sigma $ or $ \delta $):

\begin{equation}
    J(p) \coloneqq \int_{t_{\mathrm{start}}}^{t_{\mathrm{end}}} \dddot{p}(t)^{2} \,\mathrm{d}t.
\end{equation}

In Equations \ref{eq:frenet_cost_lat} to \ref{eq:frenet_cost_col}, the subscripts $ \mathrm{start} $ and $ \mathrm{end} $ respectively refer to the current position of the ship and the position up to which planning is performed. $ \dot{\sigma}_{\mathrm{target}} $ denotes the target longitudinal speed. The variable $ t $ is the time with the current ship position as origin. $ D_T $ denotes the lateral distance between the currently considered trajectory and a colliding trajectory labeled $ T $.
$ K_{\mathrm{J}} $, $ K_{\mathrm{t}} $, $ K_{\mathrm{\delta}} $, $ K_{\mathrm{\sigma}} $ and $ K_{\mathrm{D}} $ are adjustable parameters.

In our implementation, we added the $ C_{\mathrm{col}} $ term in the cost function, which is an extension with respect to the original PythonRobotics library: this term is used to assign a higher cost to trajectories close to colliding trajectories and a lower cost to trajectories further from colliding trajectories.
This extension provides a more realistic avoidance by increasing the distance to obstacles, as we verified with various obstacle shapes in our simulations.

\section{Experiments}
\label{sec:experiments}

In order to evaluate the proposed methods we perform experiments in two different scenarios. In the architecture in Figure \ref{fig:architecture}, we replace the ship control, perception, localisation and fusion components with a simulation of the ship. We use the simulation environment explained in Section \ref{sec:simulation_env}. To be able to deal with the waypoints that our algorithm provides, we use a line of sight path following algorithm. This controls the desired heading of the ship in order to sail in a straight line toward the next waypoint. 
For the occupancy grid map we use images with pixels with a size of 3.125m. Both the PPO agent and MPRL use input images with a resolution of 64x64 pixels. This means that the observation contains information 100m in each direction from the ship. The ship we use in our simulations has a length of 15m and a width of 4m. When detecting collisions, we add a safety margin of 2m around the ship to keep a safe distance from any obstacles. The global path planning supplies waypoints that are between 150m and 200m apart unless stated otherwise in a specific experiment.

\section{Results}
\label{sec:results}
In this section, we describe two experiments to evaluate the performance of PPO, Frenet frame navigation and MPRL. These two experiments simulate two different scenarios which can occur in real world shipping applications.

\subsection{Scenario 1: Straight Path with Obstacles}

The first scenario contains a straight waterway with a sequence of obstacles on the sides simulating a realistic scenario in which vessels are berthed along the quay walls. The paths for each of the approaches can be seen in Figure \ref{fig:results_scen1}. The first thing that is clear is that PPO is not able to navigate through this scenario. It is able to avoid the first couple obstacles very well but fails to avoid the fourth obstacle because the passage between the obstacle and the quay wall is too narrow. MPRL and Frenet frame navigation however are able to avoid the obstacles. The Frenet frame planning takes a more straight path toward the goal, passing dangerously close to some obstacles. MPRL has a path with more undulations but stays further away from obstacles which makes the path safer. 

For a quantitative comparison and to investigate this in more detail we look at the cumulative distribution function of the distance towards the nearest obstacle. This can be seen in Figure \ref{fig:CDF_scen1}. Here, we clearly see that Frenet frame based navigation results in a trajectory that regularly passes very close to an obstacle. Around 20\% of the trajectory is closer than 10m to an obstacle. For MPRL, this is only the case for around 8\% of its trajectory. This clearly shows that MPRL chooses a trajectory that is better at avoiding obstacles.

\begin{figure}[t]
    \centering
    \includegraphics[width=0.45\linewidth, angle=270,origin=c]{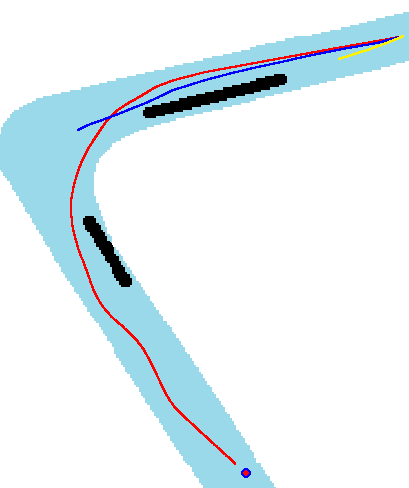}
    \caption{Paths of the different path planning methods in scenario 2 (Red - MPRL, Yellow - PPO, Blue - Frenet Frame Navigation) in which the black pixels (a one in the occupancy grid map) represent the objects on the waterway.}
    \label{fig:results_scen2}
\end{figure}

\subsection{Scenario 2: Corner with Obstacles}
The second evaluation scenario consists of a left turn containing two obstacles. Using this scenario we can evaluate how safely each of the approaches chooses to navigate through the turn. The different trajectories can be seen in Figure \ref{fig:results_scen2}. 
In this scenario, the Frenet frame and PPO baselines both fail to navigate to the goal. PPO does not manage to navigate past the first obstacle. The Frenet frame approach successfully avoids the first obstacle but cannot navigate through the corner while also avoiding the second obstacle. We also needed to modify the configuration parameters of the Frenet frame approach compared to the previous scenario to acquire these results. 
The global path planner provides global waypoints to the Frenet frame method that are between 15m and 30m apart. In every other experiment we have used a global path with waypoints between 150m and 200m apart.
This makes the use of Frenet frame planning in practice more difficult since the configuration is dependent on the situation.

MPRL is the only approach that is able to reach the goal in this scenario. We see that it can successfully avoid both obstacles. However, there are some undulations visible in the path. We hypothesize that this behaviour can be improved in the future by adding a component to the reward that is dependent on the distance of the ship to the nearest obstacle. This will encourage the agent to maximize its distance to any obstacle instead of only avoiding obstacles. Adding this reward component will also improve the behaviour of the PPO agent. MPRL could be further improved by increasing the occupancy grid map resolution which allows the agent to make more fine grained simulations and to navigate more narrow waterways. This will however increase the computational requirement during execution.


\section{Conclusion}
\label{sec:conclusion}
In this paper, we have presented a novel path planning system called Model Predictive Reinforcement Learning (MPRL). We developed a novel simulation environment which represents the environment using an occupancy grid map allowing us to deal with an unknown number of obstacles of any shape as well as any shape of waterway. We compare our approach with path planning using a Frenet frame and an approach based on a PPO agent. We evaluate each of these methods on two different scenarios. Our results showed that PPO is the least capable of navigating through narrow waterways containing obstacles. It was not able to reach the goal in any of the tested scenarios. Frenet frame planning was able to navigate to the goal in a straight waterway with obstacles but had trouble with a corner containing obstacles. We also determined that practical use of the Frenet frame algorithm is difficult since the configuration is dependent on the situation. MPRL managed to navigate to the goal in both test scenarios, taking a safe path away from obstacles in both cases. 


\section*{Acknowledgements}
The imec icon Smart Waterway project runs from October 1st 2019 until February 28th 2022 and combines the expertise of industrial partners Seafar, Pozyx, Citymesh and Blue Line Logistics with the scientific expertise of imec research partners IDLab (University of Antwerp and University of Ghent) and TPR from University of Antwerp. The project was realised with the financial support of Flanders Innovation \& Entrepreneurship (VLAIO, project no. HBC.2019.0058).
Astrid Vanneste and Simon Vanneste are supported by the Research Foundation Flanders (FWO) under Grant Number 1S12121N and Grant Number 1S94120N respectively.
We would like to thank Aleksander Chernyavskiy (Seafar NV) for the fruitful discussions about the design of the applications presented in this paper and for allowing us to carry out tests on Seafar's simulation system.
We are grateful to Ahmed Ahmed (IDLab) for the help he provided in reviewing classical planning algorithms.

\bibliographystyle{IEEEtran}
\bibliography{references}

\end{document}